%% file: paper.tex
\documentclass[11pt,letterpaper]{article}
\usepackage{acl2016}
\usepackage{subcaption}
\usepackage{times}
\usepackage{url}
\usepackage{latexsym}
\usepackage{graphicx}
\usepackage{amsmath}
\usepackage{dsfont}
\usepackage{float}
\usepackage{multirow}
\usepackage{url}
\usepackage{comment}
\usepackage[usenames,dvipsnames,table]{xcolor}
\usepackage{epsfig}
\usepackage{multirow}
\usepackage{microtype}
\usepackage{amsfonts}
\usepackage{amssymb}
\usepackage{amsthm}
\usepackage{graphicx}
\usepackage{verbatim}
\usepackage{hhline}
\usepackage{booktabs}
\usepackage{tabularx}
\usepackage{balance}

\usepackage[table]{xcolor}
\makeatletter
\newcommand{\@BIBLABEL}{\@emptybiblabel}
\newcommand{\@emptybiblabel}[1]{}
\makeatother
\allowdisplaybreaks

\def\aclpaperid{797}
\aclfinalcopy

\newcommand{\cut}[1]{}

\newcommand{\tightlist}{\setlength{\parskip}{0ex}\setlength{\itemsep}{0ex}\setlength{\parsep}{0ex}}

\newcommand{\para}[1]{\noindent\textbf{#1} }

\newcommand{\hannahedit}[1]{#1}

\title{Connotation Frames: A Data-Driven Investigation}
\author{
Hannah Rashkin \:  \: \: \: Sameer Singh \:  \: \: \: Yejin Choi \\
Computer Science \& Engineering \\
University of Washington \\
      {\tt \{hrashkin, sameer, yejin\}@cs.washington.edu} 
}

\date{}

\begin{document}
\maketitle
\input{abstract}

\input{introduction}
\input{connframe}
\input{graph}
\input{lbp}
\input{results}
\input{analysis}
\balance
\input{related}
\input{conclusions}

\section*{Acknowledgements}
We thank the anonymous reviewers for many insightful comments. We also thank members of UW NLP for discussions and support.  
This material is based upon work supported by the National Science Foundation Graduate Research Fellowship Program under Grant No. DGE-1256082, in part by NSF grants IIS-1408287, IIS-1524371, and gifts by Google and Facebook. 

\bibliographystyle{acl2016}
\bibliography{paper}


\end{document}

%% file: abstract.tex

\begin{abstract}
\hannahedit{
Through a particular choice of 
a predicate (e.g., ``$x$ violated $y$''), a writer can subtly connote 
a range of implied sentiments and presupposed facts 
about the entities $x$ and $y$:} 
(1) \emph{writer's perspective}: projecting $x$ as an ``antagonist'' and $y$ as a ``victim'', 
(2) \emph{entities' perspective}: $y$ probably dislikes $x$, 
(3) \emph{effect}: something bad happened to $y$, 
(4) \emph{value}: $y$ is something valuable, and
(5) \emph{mental state}: $y$ is distressed by\cut{most likely not happy about} the event.

We introduce \emph{connotation frames} as a representation formalism to organize these rich dimensions of 
connotation 
using typed relations.  
First, we investigate the feasibility of obtaining 
connotative labels through crowdsourcing experiments. We then present models 
for predicting 
the connotation frames of 
verb predicates 
based on their distributional word representations and the interplay between different types of connotative relations. 
Empirical results confirm that connotation frames can be induced from various data sources that reflect how people use language and give rise to the connotative meanings. We conclude with analytical results that show the potential use of connotation frames for analyzing subtle biases in online news media. 
\end{abstract}


%% file: introduction.tex

\section{Introduction}
\label{sec:intro}

\begin{figure}[t!]
	\centering   
\vspace*{-2mm}
\hspace*{-3mm}
  \includegraphics[width=1.0\columnwidth]{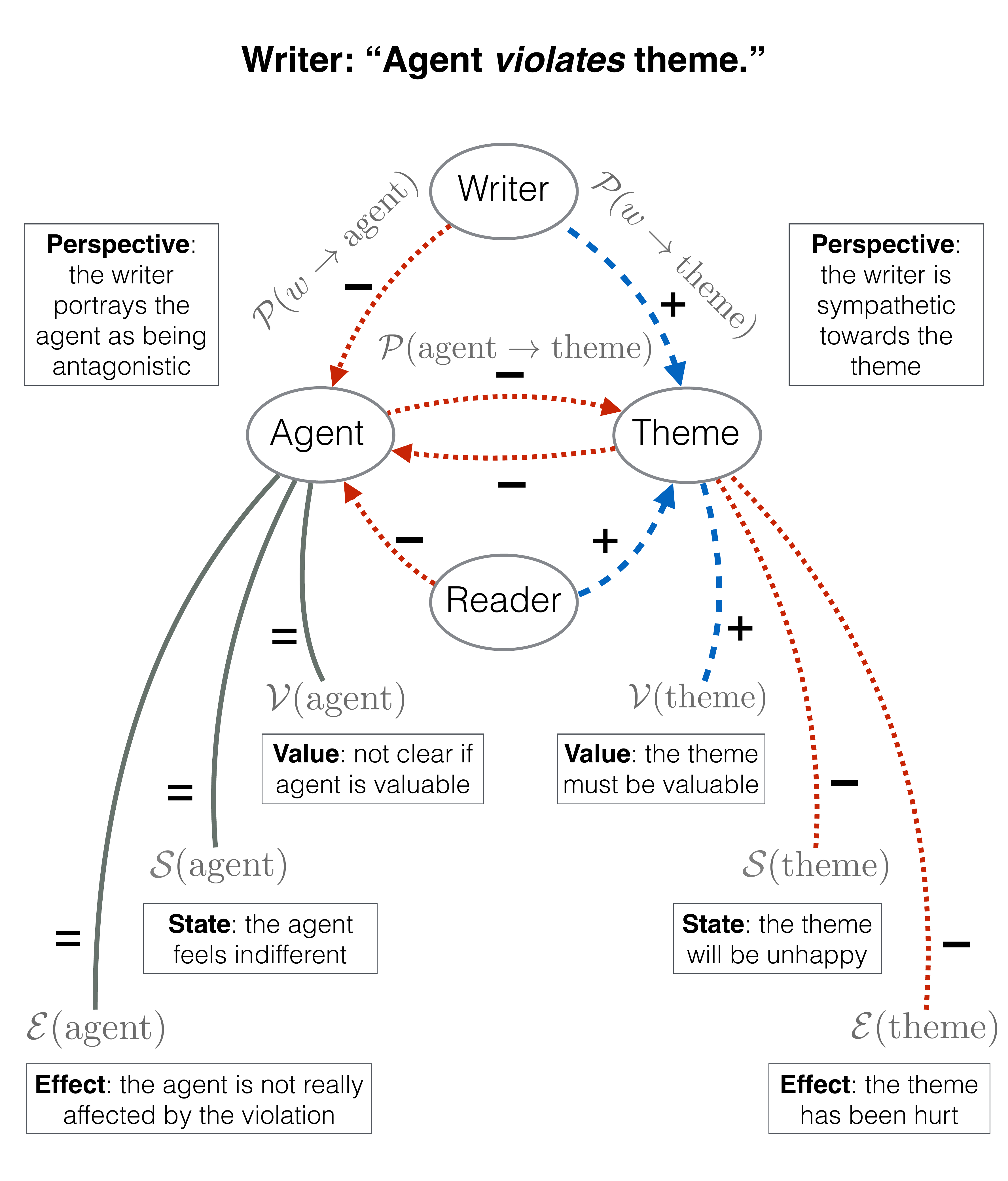}
  \vspace*{-3mm}
  \caption{
  \small 
  An example connotation frame of ``violate'' as a set of typed relations: perspective $\mathcal{P}(x \rightarrow y)$, 
  effect $\mathcal{E}(x)$,
  value $\mathcal{V}(x)$,
  and mental state $\mathcal{S}(x)$.
  }
  \label{fig:opening}    
  \vspace*{-3mm}
\end{figure}

People commonly express their opinions through 
subtle and nuanced language~\cite{ThomasPangLee,somasundaranwiebe2010}. Often, through seemingly objective statements, the writer can influence the readers' judgments 
toward an event and their participants. 
Even by choosing a particular predicate, the writer can indicate rich connotative information about the entities that interact through the predicate. Specifically, through a simple statement such as  ``$x$ violated $y$'', the writer can convey:
\vspace*{-2mm}
\begin{description}
\vspace*{-5mm}
\tightlist
\item[(1) writer's perspective:] the writer is projecting $x$ as an ``antagonist'' and $y$ as a ``victim'', eliciting negative perspective from readers toward $x$ (i.e., blaming $x$) and positive perspective toward $y$ (supportive or sympathetic to $y$).
\item[(2) entities' perspective:] $y$ most likely feels negatively toward $x$ as a result of being violated.
\item[(3) effect:] something bad happened to $y$.  
\item[(4) value:] $y$ is something valuable, since it does not make sense to violate something worthless. In other words, the writer is presupposing $y$'s positive value as a fact.
\item[(5) mental state:] $y$ is most likely unhappy about the outcome.\footnote{To be more precise, $y$ is most likely in a negative mental state assuming it is an entity that can have a mental state.} 
\end{description}

\begin{table*}[tb]
\begin{center}
\footnotesize
\begin{tabular}{lp{1.25in}p{.85in}p{2.375in}l}
\toprule
\bf Verb &  \multicolumn{2}{c}{\bf Subset of Typed Relations} & \bf Example Sentences &\bf L/R \\
\midrule
suffer& 
$\mathcal{P}(w \rightarrow \mbox{{agent}}) = + 
\newline 
\mathcal{P}(w \rightarrow \mbox{{theme}}) = - 
\newline 
\mathcal{P}(\mbox{{agent}}\rightarrow\mbox{{theme}}) = - $
&
$\mathcal{E}(\mbox{{agent}}) = -
\newline 
\mathcal{V}(\mbox{{agent}}) = +
\newline 
\mathcal{S}(\mbox{{agent}}) = -
$
& The story begins in Illinois in 1987, when \emph{a 17-year-old girl} \textbf{suffered} \emph{a botched abortion.} &R\\

\midrule
guard & 
$\mathcal{P}(w\rightarrow \mbox{{agent}}) = + 
\newline 
\mathcal{P}(w\rightarrow\mbox{{theme}}) = +
\newline 
\mathcal{P}(\mbox{{agent}} \rightarrow \mbox{{theme}}) = + $
&
$\mathcal{E}(\mbox{{theme}}) = +
\newline 
\mathcal{V}(\mbox{{theme}}) = +
\newline 
\mathcal{S}(\mbox{{theme}}) = +
$
&In August, \emph{marshals} \textbf{guarded} \emph{25 clinics} in 18 cities.
& L
\\
\midrule
uphold 
&
$\mathcal{P}(w\rightarrow\mbox{{theme}}) = + 
\newline 
\mathcal{P}(\mbox{{agent}} \rightarrow \mbox{{theme}}) = + $
&
$\mathcal{E}(\mbox{{theme}}) = +
\newline 
\mathcal{V}(\mbox{{theme}}) = +
$
&
{A hearing is scheduled 
to make a decision on 
whether to \textbf{uphold} \emph{the clinic's suspension}.}
&
R
\\
\bottomrule
\end{tabular}\end{center}
\vspace*{-3mm}
\caption{\small Example typed relations 
(perspective $\mathcal{P}(x \rightarrow y)$, 
  effect $\mathcal{E}(x)$,
  value $\mathcal{V}(x)$,
  and mental state $\mathcal{S}(x)$), where $w$ denotes the writer. 
Not all typed relations are shown due to space constraints. 
The example sentences demonstrate the usage of the predicates in left [L] or right [R] leaning news sources.
}
\label{table:news-bias-examples}
\end{table*}

Even though the writer might not explicitly state any of the interpretations [1-5] above, the readers will be able interpret these intentions as a part of their comprehension. In this paper, we present an empirical study of how to represent and induce the connotative interpretations that can be drawn from a verb predicate, as illustrated above. 

We introduce \emph{connotation frames} as a representation framework to organize the rich dimensions of the implied sentiment and presupposed facts. Figure~\ref{fig:opening} shows an example of a connotation frame for the predicate \emph{violate}. We define four different typed relations: $\mathcal{P}(x \rightarrow y)$ for perspective of $x$ towards $y$, $\mathcal{E}(x)$ for effect on $x$, $\mathcal{V}(x)$ for value of $x$, and $\mathcal{S}(x)$ for mental state of $x$.  These relationships can all be either positive (+), neutral (=), or negative (-).

Our work is the first study to investigate frames as a representation formalism for connotative meanings. 
This contrasts with previous computational studies and resource development for frame semantics,  where the primary focus was almost exclusively on denotational meanings of language~\cite{baker1998berkeley,palmer2005proposition}. Our formalism draws inspirations from the earlier work of frame semantics, however, in that we investigate the connection between a word and the related world knowledge associated with the word~\cite{chuck76}, which is essential for the readers to interpret 
many layers of the implied sentiment and presupposed value judgments.

We also build upon the extensive amount of literature in sentiment analysis~\cite{pang2008,liu2012survey}, especially the recent emerging efforts on implied sentiment analysis~\cite{feng2013connotation,Greene:2009:MWS:1620754.1620827}, entity-entity sentiment inference~\cite{DBLP:journals/corr/WiebeD14}, opinion role induction~\cite{wiegand2015}, and effect analysis~\cite{choi-wiebe:2014:EMNLP2014}. However, our work is the first to organize 
aspects of the connotative information into coherent frames. 

More concretely, our contributions are threefold: (1) a new formalism, model, and annotated dataset for studying the connotation frames from large-scale natural language data and statistics, (2) data-driven insights into the dynamics among different typed relations within each frame, and (3) an analytic study to show the potential use of connotation frames for analyzing subtle biases in journalism. 

The rest of the paper is organized as follows: in \S2, we provide the  definitions and data-driven insights for connotation frames. In \S3, we introduce models for inducing the connotation frames, followed by empirical results, annotation studies, and analysis on news media in \S4. We discuss related work in \S5 and conclude in \S6.

%% file: connframe.tex
\section{Connotation Frame}
\label{sec:frame}

\input{table}

Given a predicate $v$, we define a connotation frame $\mathcal{F}(v)$ as a collection of typed relations and their polarity assignments: 
(i) \textbf{perspective} $\mathcal{P}^v(x_i\rightarrow x_j)$: whether the predicate $v$ implies  directed sentiment from the entity $x_i$ to the entity $x_j$, (ii) \textbf{value} $\mathcal{V}^v(x_i)$: whether $x_i$ is presupposed to be valuable by the predicate $v$, (iii) \textbf{effect} $\mathcal{E}^v(x_i)$: whether the event denoted by the predicate $v$ is good or bad for the entity $x_i$, and (iv) \textbf{mental state} $\mathcal{S}^v(x_i)$: the likely mental state of the entity $x_i$ as a result of the event. 
We assume that each typed relation can have one of the three connotative polarities $\in \{+,-,=\}$, i.e., positive, negative, or neutral. 
Our goal in this paper is to focus on the general connotation of the predicate considered out of context. We leave contextual interpretation of connotation as future work. 

Table~\ref{table:news-bias-examples} shows examples of connotation frame relations for the verbs \emph{suffer}, \emph{guard}, and \emph{uphold}, along with example sentences. 
For instance, for the verb \emph{suffer}, 
the writer is likely to have a positive perspective towards the agent (e.g., being supportive or sympathetic toward the ``17-year-old girl'' in the example shown on the right) and a negative perspective towards the theme (e.g., being negative towards `botched abortion''). 

\subsection{Data-driven Motivation}

Since the meaning of language is ultimately contextual, the exact connotation will vary depending on the context of each utterance. Nonetheless, there still are common shifts or biases in the connotative polarities, as we found from two data-driven analyses.  

First, we looked at words from the Subjectivity Lexicon~\cite{wilson2005recognizing} that are used in the argument positions of a small selection of predicates in Google Syntactic N-grams~\cite{goldberg-orwant:2013:*SEM}.  \hannahedit{For this analysis, we assumed that the agent is the word in the subject position while the theme is the word in the object position. We found 64\% of the words in the agent role of \emph{suffer} are positive, and 94\% of the words in the theme role are negative, which is consistent with the polarities of the writer's perspective towards these arguments, as shown in Table~\ref{table:news-bias-examples}.  For \emph{guard}, 57\% of the agents and 76\% of the themes are positive, and in the case of \emph{uphold}, 56\% of the agents and 72\% of the themes are positive}.

We also investigated how media bias can potentially be analyzed through connotation frames. 
%
From the Stream Corpus 2014 dataset~\cite{streamcorpus14}, we selected  all articles 
from news outlets with known political biases,\footnote{The articles come from $30$ news sources indicated by others as exhibiting liberal or conservative leanings~\cite{sourcelist-1,sourcelist-2,sourcelist-3,sourcelist-4}} 
and compared how they use polarised words such as ``accuse'', ``attack'', and ``criticize'' differently in light of $\mathcal{P}(w \rightarrow agent)$ and $\mathcal{P}(w \rightarrow theme)$ relations of the connotation frames.  
%
Table~\ref{tab:analysis:verbs} shows interesting contrasts.  
{Obama}, for example, is frequently portrayed as someone who \emph{attacks}
or \emph{criticizes} others according to the right-leaning sources, whereas the left-leaning sources portray Obama as the victim of harsh acts like ``attack'' or ``criticize''.\footnote{That is, even if someone truly deserves criticism from Obama, left-learning sources would choose slightly different wordings to avoid a potentially harsh portrayal of Obama.}
%
Furthermore, by knowing the perspective relationships $\mathcal{P}(w \rightarrow x_i)$ associated with a predicate, we can  make predictions about how the left-leaning and right-leaning sources feel about specific people or issues.  For example, because left-leaning sources frequently use McCain, Trump, and Limbaugh in the agent position of attack, we might predict that these sources have a negative sentiment towards these entities.

\begin{table*}[tb]
\begin{center}
{\footnotesize
\setlength{\tabcolsep}{23pt}
\begin{tabularx}{\textwidth}{XcX}
\toprule
\multicolumn{3}{c}{\parbox[t]{.9\textwidth}{{\bf Perspective Triad:} If argument $x_i$ is positive towards $x_j$, and $x_j$ is positive towards $x_k$, then we expect $x_i$ is also positive towards $x_k$.  
Similar dynamics  hold for the negative case.}}\\
&$\mathcal{P}_{w \rightarrow x_i} = \neg \ ( \mathcal{P}_{w \rightarrow x_j} \oplus \mathcal{P}_{x_i \rightarrow x_j} ) $ &\\ 

\midrule
\multicolumn{3}{c}{\parbox[t]{.9\textwidth}{{\bf Perspective -- Effect:} If a predicate has a positive effect on one argument, then we expect that the interaction between the arguments was positive. 
Similar dynamics  hold for the negative case. 
}}\\
&$\mathcal{E}_{x_i} = \mathcal{P}_{x_j \rightarrow x_i}$ &\\

\midrule
\multicolumn{3}{c}{\parbox[t]{.9\textwidth}{{\bf Perspective -- Value:} If argument $x_i$ is presupposed as valuable, then we expect that the writer also views $x_i$ positively.  Similar dynamics hold for the negative case.}}\\
&$\mathcal{V}_{x_i} = \mathcal{P}_{w \rightarrow x_i}$ &\\

\midrule
\multicolumn{3}{c}{\parbox[t]{.9\textwidth}{{\bf Effect -- Mental State:} If the predicate has a positive effect on an argument $x_i$, then we expect that $x_i$ will gain a positive mental state. Similar dynamics  hold for the negative case.}}\\
&$\mathcal{S}_{x_i} = \mathcal{E}_{x_i}$ &\\

\bottomrule
\end{tabularx}}
\end{center}
\vspace*{-3mm}
\caption{\small Potential Dynamics among Typed Relations: 
we propose models that parameterize these dynamics using log-linear models (frame-level model in \S3).
}
\vspace*{-4mm}
\label{table:infrules}
\end{table*}

\subsection{Dynamics between Typed Relations} 
\label{sec:frame:inf}
Given a predicate, the polarity assignments of typed relations are interdependent. 
For example, if the writer feels positively towards the agent but negatively towards the theme, then it is likely that the agent and the theme do not feel positively towards each other. 
This insight is related to that of \newcite{DBLP:journals/corr/WiebeD14}, but differs in that the polarities are predicate-specific and do not rely on knowledge of prior sentiment towards the arguments, themselves. 
This and other possible interdependencies are summarized in Table~\ref{table:infrules}.  These interdependencies serve as general guidelines of what properties we expect to depend on one another, especially in the case where the polarities are non-neutral. 
We will promote these internal consistencies in our factor graph model~(\S\ref{sec:model}) as soft constraints. 

There also exist other interdependencies that we will use to simplify our task.
First, the directed sentiments between the agent and the theme are likely to be reciprocal, or at least do not directly conflict with $+$ and $-$ simultaneously. This intuition follows from a notion of balance derived by social theory \cite{social}. Therefore, we assume that  $\mathcal{P}(x_i \rightarrow x_j)=\mathcal{P}(x_j \rightarrow x_i) = \mathcal{P}(x_i \leftrightarrow x_j)$, and we only measure for these binary relationships going in one direction. In addition, we assume the \emph{predicted}\footnote{Surely different readers can and will form varying opinions after reading the same text. Here we concern with the most likely perspective of the general audience, as a result of reading the text.} perspective from the reader $r$ to an argument, $\mathcal{P}(r \rightarrow x)$, is likely to be the same as the \emph{implied} perspective from the writer $w$ to the same argument, $\mathcal{P}(w \rightarrow x)$. So, we only try to learn the perspective of the writer. Lifting these assumptions will be future work. 

For simplicity, our work only explores verb predicates and focuses on the polarities involving the agent and the theme roles, which we will refer to as $a$ and $ t$.  We will assume that these roles are correlated to the subject and object positions. 

%% file: table.tex
\begin{table*}[tb]
\begin{center}
{\small
\begin{tabular}{ccccc}
\toprule
\bf Verb & \bf $x$'s role & $\mathbf{\mathcal{P}(w\rightarrow\cdot)}$ & \bf Left-leaning Sources & \bf Right-leaning Sources \\
\midrule
\multirow{2}{*}{\em accuse} & \cellcolor{lightgray!35}{agent} &\cellcolor{lightgray!35}{-}
& \cellcolor{lightgray!35}{Putin, Progressives, Limbaugh, Gingrich  }
& \cellcolor{lightgray!35}{ activist, U.S., protestor, Chavez} \\
						& \cellcolor{lightgray!10}{theme }& \cellcolor{lightgray!10}{ +}
& \cellcolor{lightgray!10}{official, rival, administration, leader }
& \cellcolor{lightgray!10}{Romney, Iran, Gingrich, regime } \\
 					    \midrule
\multirow{2}{*}{\em attack} &\cellcolor{lightgray!35}{agent} & \cellcolor{lightgray!35}{-}
&\cellcolor{lightgray!35}{ McCain, Trump, Limbaugh   }	  
& \cellcolor{lightgray!35}{\textbf{Obama}, campaign, Biden, Israel} \\
 	&\cellcolor{lightgray!10}{theme} &\cellcolor{lightgray!10} {+}
& \cellcolor{lightgray!10}{Gingrich, \textbf{Obama}, policy }  
& \cellcolor{lightgray!10}{citizen, Zimmerman }  \\
 					    \midrule
\multirow{2}{*}{\em criticize} & \cellcolor{lightgray!35}{agent} & \cellcolor{lightgray!35}{-}
& \cellcolor{lightgray!35}{ Ugandans, rival, Romney, Tyson  }
&  \cellcolor{lightgray!35}{ Britain, passage, \textbf{Obama}, Maddow }\\
 					       &\cellcolor{lightgray!10}{theme} & \cellcolor{lightgray!10}{+}
& \cellcolor{lightgray!10}{ \textbf{Obama}, Allen, Cameron, Congress } 
& \cellcolor{lightgray!10}{ Pelosi, Romey, GOP, Republicans}   \\
\bottomrule
\end{tabular}
}
\end{center}
\vspace*{-3mm}
\caption{ \small
Media Bias in Connotation Frames: {Obama}, for example, is portrayed as someone who \emph{attacks} or \emph{criticizes} others by the right-leaning sources, whereas the left-leaning sources portray Obama as the victim of harsh acts like ``attack'' and ``criticize''.}
\label{tab:analysis:verbs}
\end{table*}

\cut{
}

%% file: graph.tex
\begin{figure}[t]\centering
\vspace*{-1mm}
    \includegraphics[width=0.48\textwidth]{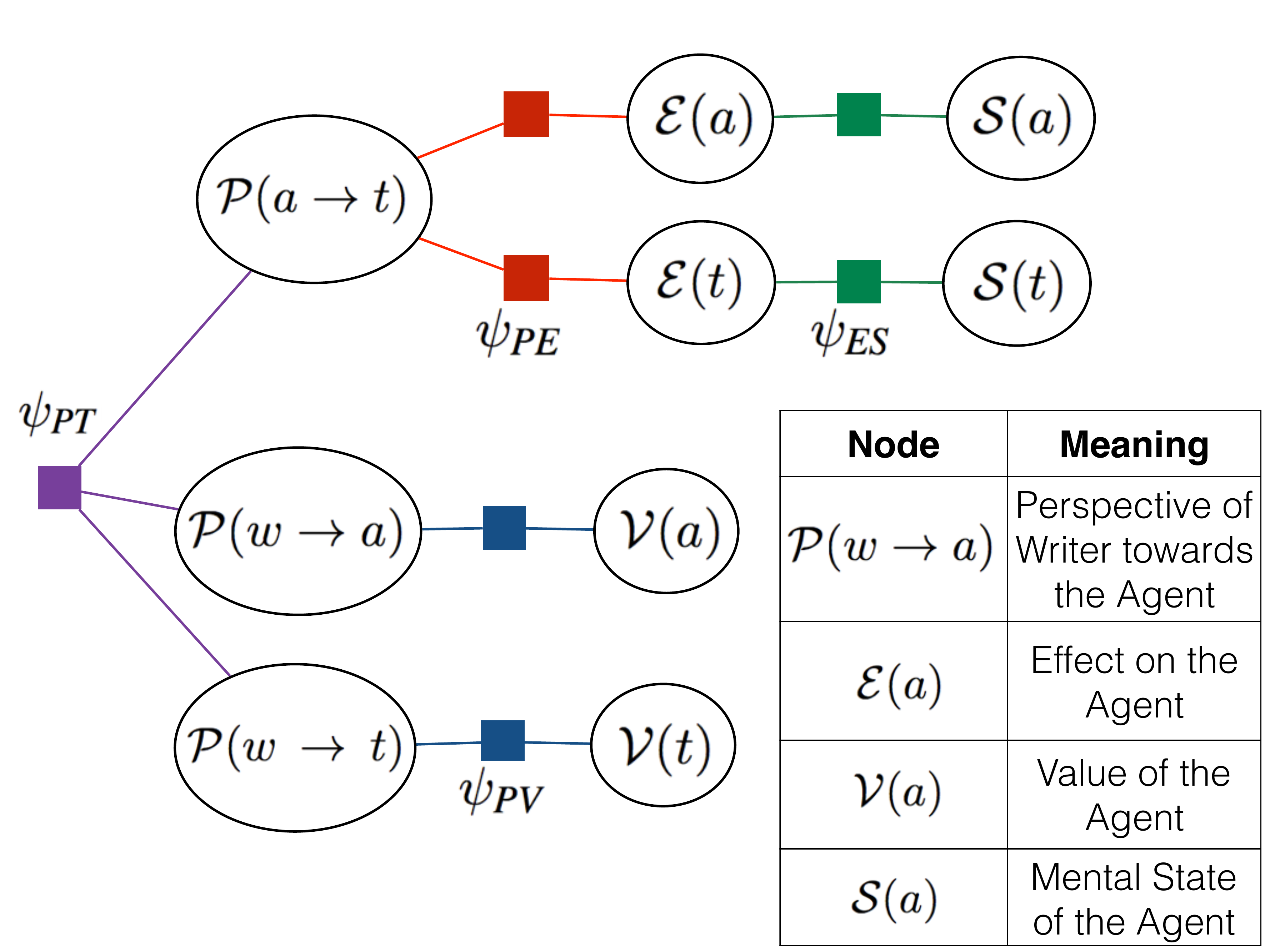}
    \vspace*{-3mm}
  \caption{ \small A factor graph for predicting 
  the polarities of the typed relations that define a connotation frame for a given verb predicate. 
  The factor graph also includes unary factors ($\psi_{emb}$), which we left out for brevity. 
  }
  \label{graphpic} 
\end{figure}
\section{Modeling Connotation Frames}
\label{sec:model}

Our task is essentially that of lexicon induction~\cite{Akkaya:2009:SWS:1699510.1699535,feng2013connotation} 
in that we want to induce the connotation frames of previously unseen verbs. For each verb predicate, we infer a connotation frame composed of 9 relationship aspects that represent: \emph{perspective} 
$\{\mathcal{P}(w\rightarrow t)$,
$\mathcal{P}(w\rightarrow a)$,
$\mathcal{P}(a\rightarrow t)\}$, \emph{effect}
$\{\mathcal{E}(t)$,
$\mathcal{E}(a)\}$, \emph{value} 
$\{\mathcal{V}(t)$,
$\mathcal{V}(a)\}$, and \emph{mental state}
$\{\mathcal{S}(t)$,
$\mathcal{S}(a)\}$  polarities,
where $w, a, t$ denote the writer, the agent, and the theme, respectively. 

We propose two models: an aspect-level model that makes the prediction for each typed relation independently based on the distributional representation of the context in which the predicate commonly appears (\S\ref{ssec:aspect}), and a frame-level model that makes the prediction over the connotation frame collectively in consideration of the dynamics between typed relations (\S\ref{ssec:frame}). 

\subsection{Aspect-Level}
\label{ssec:aspect}
  
Our aspect-level model predicts labels for each of these typed relations separately.  As input, we use the 300-dimensional dependency-based word embeddings from \newcite{levy:depemb}. For each aspect, there is a separate MaxEnt (maximum entropy) classifier used to predict the label of that aspect on a given word-embedding, which is treated as a 300 dimensional input vector to the classifier.  The MaxEnt classifiers learn their weights using LBFGS on the training data examples with re-weighting of samples to maximize for the best average F1 score.

\subsection{Frame-Level}
\label{ssec:frame}

Next we present a factor graph model (Figure~\ref{graphpic}) of the connotation frames that parameterizes the dynamics between typed relations. 
Specifically, for each verb predicate, 
the factor graph contains 9 nodes representing the different aspects of the connotation frame involving the writer ($w$), the agent ($a$), and the theme ($t$).
All these variables take polarity values from the set $\{-, =, +\}$.


We define $\mathbf{Y}_i := \{\mathcal{P}_{wt}, \allowbreak\mathcal{P}_{wa}, \allowbreak\mathcal{P}_{at}, \allowbreak\mathcal{E}_{t},  \allowbreak\mathcal{E}_{a}, \mathcal{V}_{t}, \allowbreak\mathcal{V}_{a}, \allowbreak\mathcal{S}_{t}, \mathcal{S}_{a} \}$ as the set of relational aspects for the $i^{th}$ verb predicate.  The factor graph for $\mathbf{Y}_i$, is illustrated in Figure~\ref{graphpic}, and we describe the factor potentials, $\psi$, in detail in the rest of this section.  The probability of an assignment of polarities to the nodes in $\mathbf{Y}_i$ is:

\vspace*{-4mm}
\begin{align*}
P(\mathbf{Y}_i) ~\propto
&~~~ \psi_\text{PV}({\mathcal{P}_{wa}},{\mathcal{V}_a})~\psi_\text{PV}({\mathcal{P}_{wt}},{\mathcal{V}_{t}})\nonumber\\
&~~~\psi_{\text{PE}}({\mathcal{P}_{at}},{\mathcal{E}_a})~\psi_{\text{PE}}({\mathcal{P}_{at}},{\mathcal{E}_t})\nonumber\\
&~~~\psi_{\text{ES}}({\mathcal{E}_a},{\mathcal{S}_a})~\psi_{\text{ES}}({\mathcal{E}_t},{\mathcal{S}_t})\nonumber\\&~~~\psi_\text{\text{PT}}({\mathcal{P}_{wt}},{\mathcal{P}_{wa}},{\mathcal{P}_{at}})\prod_{y \in \mathbf{Y}_i}\psi_\text{\emph{emb}}(y)\nonumber\\
\end{align*} 

\vspace*{-6mm}
\para{Embedding Factors}
We include 
{unary} factors on all nodes to represent the results of the aspect-level classifier. 
Incorporating this knowledge as factors, as opposed to \emph{fixing} the variables as observed, affords us the flexibility of representing noise in the labels as soft evidence.  The potential function $\psi_\text{\emph{emb}}$ is a log-linear function of a feature vector f, which is a one-hot feature vector representing the polarity of a node ($+$,$-$,or $=$). For example, with the node representing the value of the theme ($\mathcal{V}_t$):
\[
\psi_{emb}(\mathcal{V}_t) = e^{\theta_{\mathcal{V}_t}\cdot f(\mathcal{V}_t)}
\]
The potential $\psi_{emb}$ is defined similarly for the remaining eight nodes.

Weights $\theta$ are learned in a piecewise likelihood manner~\cite{Sutton2009} for each factor independently using stochastic gradient descent (SGD) over the training data.

\paragraph{Interdependency Factors}
We include {interdependency} factors to promote the properties defined by the dynamics between relations (\S\ref{sec:frame:inf}).  The potentials for Perspective Triad, Perspective-Value, Perspective-Effect, and Effect-State Relationships ($\psi_\text{\emph{PT}}$, $\psi_\text{\emph{PV}}$,  $\psi_\text{\emph{PE}}$, $\psi_\text{\emph{ES}}$ respectively) are all defined using log-linear functions of one-hot feature vectors that encode the combination of polarities of the neighboring nodes.  Thus the potential $\psi_\text{\emph{PT}}$ is:
\[
\psi_\text{\emph{PT}}({\mathcal{P}_{wt}},{\mathcal{P}_{wa}},\mathcal{P}_{at}) = e^{\theta_{PT}\cdot f({\mathcal{P}_{wt}},{\mathcal{P}_{w a}},{\mathcal{P}_{at})}}
\]

\noindent
And we define the potentials for $\psi_\text{\emph{PV}}$, $\psi_\text{\emph{PE}}$, and $\psi_\text{\emph{ES}}$ for nodes pertaining to the agent as:
\[
\psi_\text{\emph{PV}}({\mathcal{P}_{wa}}, {\mathcal{V}_a}) = e^{\theta_{PV,a}\cdot f({\mathcal{P}_{wa}},{\mathcal{V}_a})}
\]
\[
\psi_\text{\emph{PE}}({\mathcal{P}_{at}},{\mathcal{E}_a}) = e^{\theta_{PE,a}\cdot f({\mathcal{P}_{at}},{\mathcal{E}_a})}
\]
\[
\psi_\text{\emph{ES}}({\mathcal{E}_a},{\mathcal{S}_a}) = e^{\theta_{ES,a}\cdot f({\mathcal{E}_a},{\mathcal{S}_a})}
\]
and we define the potentials for the theme nodes similarly. 
As with the unary seed factors, weights $\theta$ are learned using SGD over training data.

%% file: lbp.tex
\paragraph{Belief Propagation}
We use belief propagation to induce the connotation frames of previously unseen verbs. 
In the belief propagation algorithm, messages are iteratively passed between the nodes to their neighboring factors.  
Each message $\mu$, containing a scalar for each value $x\in \{-,=,+\}$, is defined from each node $v$ to a neighboring factor $f$ as follows:
\vspace*{-2mm}
\[\mu_{v\rightarrow f}(x) \propto \prod_{f^*\in N(v)\setminus a}\mu_{f^*\rightarrow v}(x)\]
\noindent
 and from each factor $a$ to a neighboring node $v$ as:
\[\mu_{f\rightarrow v} \propto \sum_{x', x'_v = x} \psi(x') \prod_{v^*\in N(f)\setminus v}\mu_{v^*\rightarrow f}(x'_{v^*})\]
\noindent
 Our factor graph does not contain any loops, so we are able to perform exact inference by choosing a root node and performing message passing from the leaves to the root and back to the leaves.
At the conclusion of message passing, 
the probability of a specific polarity associated with node $v$ being equal to $x$ is proportional to $ \prod_{f \in N(v)} \mu_{f\rightarrow v}(x)$. 



%% file: results.tex

\section{Experiments}
\label{sec:results}

We first describe crowd-sourced annotations (\S\ref{ssec:data}), then present the empirical results of predicting connotation frames (\S\ref{ssec:perspective}),
and conclude with qualitative analysis of a large corpus (\S\ref{ssec:analysis}). 

\subsection{Data and Crowdsourcing}
\label{ssec:data}

In order to understand how humans interpret connotation frames, we designed an Amazon Mechanical Turk (AMT) annotation study. 
We gathered a set of transitive verbs commonly used in the New York Times corpus ~\cite{sandhaus2008new}, selecting the 2400 verbs that are used more than 200 times in the corpus. 
Of these, AMT workers annotated the 1000 most frequently used verbs. 

\para{Annotation Design} 
In a pilot annotation experiment, we found that annotators have difficulty thinking about subtle connotative polarities when shown predicates without any context. Therefore, we designed the AMT task to provide a generic context as follows. We first split each verb predicate into 5 separate tasks that each gave workers a different generic sentence using the verb.  To create generic sentences, we used Google Syntactic N-grams \cite{goldberg-orwant:2013:*SEM} to come up with a frequently seen Subject-Verb-Object tuple which served as a simple three-word sentence with generic arguments.\footnote{Because Google Syntactic N-grams only provide dependency types and do not provide semantic roles, we approximate the agent and theme as the subject and the object respectively.} For each of the 5 sentences, we asked 3 annotators to answer questions like ``How do you think the agent feels about the event described in this sentence?'' In total, each verb has 15 annotations aggregated over 5 different generic sentences containing the verb. 

In order to help the annotators, some of the questions also allowed annotators to choose sentiment using additional classes for ``positive or neutral'' or ``negative or neutral'' for when they were less confident but still felt like a sentiment might exist.  When taking inter-annotator agreement, we count ``positive or neutral'' as agreeing with either ``positive'' or ``neutral'' classes.

\para{Annotator agreement} Table~\ref{table:data} shows agreements and data statistics. The non-conflicting (NC) agreement only counts opposite polarities as disagreement.\footnote{Annotators were asked yes/no questions related to Value, so this does not have a corresponding NC agreement score.}  From this study, we can see that non-expert annotators are able to see these sort of relationships based on their understanding of how language is used.  From the NC agreement, we see that annotators do not frequently choose completely opposite polarities, indicating that even when they disagree, their disagreements are based on the degree of connotations rather than the polarity itself.  \hannahedit{The average Krippendorff alpha for all of the questions posed to the workers is 0.25, indicating stronger than random agreement. Considering the subtlety of the implicit sentiments that we are asking them to annotate, it is reasonable that some annotators will pick up on more nuances than others.  Overall, the percent agreement is encouraging that the connotative relationships are visible to human annotators.}



\para{Aggregating Annotations} We aggregated over crowdsourced labels (fifteen annotations per verb) to create a polarity label for each aspect of a verb.\footnote{
We take the average to obtain scalar value between~$[-1.,1.]$ for each  aspect of a verb's connotation frame.  For simplicity, we cutoff the ranges of negative, neutral and positive polarities as $[-1,-0.25)$, $[-0.25,0.25]$ and $(0.25,1]$, respectively.
}
%
%

Final distributions of the aggregated labels are included in the right-hand columns of Table~\ref{table:data}. 
Notably, the distributions are skewed toward positive and neutral labels. 
The most skewed connotation frame aspect is the value $\mathcal{V}(x)$ which tends to be positive, especially for the agent argument.  This makes some intuitive sense since, as the agent actively causes the predicate event to occur, they most likely have some intrinsic potential to be valuable.  An example of a verb where the agent was labelled with negative value is ``contaminate''.  In the most generic case, the writer is using ``contaminate'' to frame the agent as being worthless (and even harmful) with regards to the other event participants. For example, in the sentence ``his touch contaminated the food,'' it is clear that the writer presupposes ``his touch'' to be of negative value in how it impacts the rest of the event.


\begin{table}[tb]
\begin{center}
{\small
\begin{tabular}{ccccc}
\toprule
\multirow{2}{*}{\bf Aspect}  &\multicolumn{2}{c}{\bf \% Agreement}  
 &\multicolumn{2}{c}{\bf Distribution}\\
\cmidrule(lr){2-3}
\cmidrule(lr){4-5}
  & Strict  & NC &  \% + &  \% - \\
 \midrule 
$\mathcal{P}(w \rightarrow t)$ & 75.6 & 95.6 & 36.6 &4.6 \\
$\mathcal{P}(w \rightarrow a)$& 76.1& 95.5 &  47.1 & 7.9 \\
$\mathcal{P}(a \rightarrow t)$&  70.4 & 91.9 &  45.8 & 5.0 \\
$\mathcal{E}(t) $&  52.3 & 94.6 & 50.3 & 20.24  \\
$\mathcal{E}(a) $&  53.5 & 96.5 & 45.1 & 4.7  \\
$\mathcal{V}(t) $&  65.2 & - & 78.64 & 2.7  \\
$\mathcal{V}(a) $&  71.9 & - & 90.32 & 1.4  \\
$\mathcal{S}(t) $&  79.9 & 98.0 & 12.8 & 14.5 \\
$\mathcal{S}(a) $&  70.4 & 92.5 & 50.72 & 8.6  \\
\bottomrule
\end{tabular}
}
\end{center}
\caption{\small Label Statistics: \% Agreement refers to pairwise inter-annotator agreement.  The strict agreement counts agreement over 3 classes (``positive or neutral'' was counted as agreeing with either + or neutral), while non-conflicting (NC) agreement also allows agreements between neutral and -/+ (no direct conflicts).  Distribution shows the final class distribution of -/+ labels created by averaging annotations.} 
\label{table:data}
\end{table}
\subsection{Connotation Frame Prediction}
\label{ssec:perspective}

Using the crowdsourced labels, we randomly divide the annotated verbs into training, dev, and held-out test sets of equal size (300 verbs each).  For evaluation we measure average accuracy and F1 score of induced labels for the 9 different connotation frame relationship types for which we have annotations: $\mathcal{P}(w\rightarrow t)$, $\mathcal{P}(w\rightarrow a)$, $\mathcal{P}(a \rightarrow t)$, $\mathcal{V}(t)$, $\mathcal{V}(a)$, $\mathcal{E}(t)$, $\mathcal{E}(a)$, $\mathcal{S}(t)$, and $\mathcal{S}(a)$, where $w$ refers to the writer, $a$ to the agent, and $t$ to the theme.


\para{Baselines} To show the non-trivial challenge of learning Connotation Frames, we include a simple majority-class baselines. The \textsc{Majority} classifier assigns each of the 9 relationships the label of the majority of that relationship type found in the training data.  Some of these relationships (in particular, the Value of agent/theme) have skewed distributions, so we expect this classifier to achieve a much higher accuracy than random but a much lower overall F1 score.  

\hannahedit{Additionally, we add a \textsc{Graph Prop} baseline that is comparable to algorithms like graph propagation or label propagation which are often used for (sentiment) lexicon induction \cite{lbprop}.  
We use a factor graph with nodes representing the polarity of each typed relation for each verb.  Binary factors connect nodes representing a particular type of relation for two similar verbs (e.g. $\mathcal{P}(w\rightarrow t)$ for verbs \emph{persuade} and \emph{convince}). These binary factors have hand-tuned potentials that are proportional to the cosine similarity of the verbs' embeddings, encouraging similar verbs to have the same polarity for the various relational aspects. We use words in the training data as the seed set and use loopy belief propagation to propagate polarities from known nodes to the unknown relationships.}

Finally, we use a \textsc{3-Nearest Neighbor} baseline that labels relationships for a verb based on the predicate's 300-dimensional word embedding representation, using the same embeddings as in our aspect-level.  \textsc{3-Nearest Neighbor} labels each verb using the polarities of the three closest verbs found in the training set.  The most similar verbs are determined using the cosine similarity between word embeddings.
%

\begin{table}[tb]
    \centering
    {\small
    \begin{tabular}{clcc}
    \toprule
    {\bf Aspect} & {\bf Algorithm} & {\bf Acc.}  & {\bf Avg F$_1$}\\ 
     \midrule
     
    \multirow{5}{*}{$\mathcal{P}(w\rightarrow t)$}&Majority& $56.52$& $24.07$\\
    & Graph Prop & 59.53  & 50.20\\
    &3-nn& 62.88&  47.93\\
    &Aspect-Level& 67.56&  56.18\\
    & Frame-Level & 67.56 &  56.18\\
     \midrule
    \multirow{5}{*}{$\mathcal{P}(w\rightarrow a)$}
    &Majority& 49.83& 22.17\\
    & Graph Prop & 52.84  & 42.93\\
    &3-nn& 55.18&  45.88\\
    &Aspect-Level& 60.54&  60.72\\
    & Frame-Level & 61.87 &  63.07\\
     \midrule
     \multirow{5}{*}{$\mathcal{P}(a\rightarrow t)$}
     &Majority& 49.83& 22.17\\
    & Graph Prop & 52.17  & 46.57\\
    &3-nn& 56.52&  52.94\\
    &Aspect-Level& 63.21&  61.70\\
    & Frame-Level & 63.88 & 62.56\\
     \midrule
     \multirow{5}{*}{$\mathcal{E}(t)$}
     &Majority& 48.83&  21.87\\
    & Graph Prop & 54.85  & 51.40\\
    &3-nn& 55.18&  51.53\\
    &Aspect-Level& 64.21& 63.63\\
    & Frame-Level & 65.22 & 64.67\\
     \midrule
     \multirow{5}{*}{$\mathcal{E}(a)$}
     &Majority& 49.83& 22.17\\
    & Graph Prop & 52.17 & 35.56\\
    &3-nn& 54.85&  42.63\\
    &Aspect-Level& 62.54&  53.82\\
    & Frame-Level & 63.88  & 56.81\\
     \midrule
     \multirow{5}{*}{$\mathcal{V}(t)$}
     &Majority& 79.60& 29.55\\
    & Graph Prop & 71.91 & 35.10\\
    &3-nn& 76.25&39.09\\
    &Aspect-Level& 75.92&  45.45\\
    & Frame-Level & 76.25 & 48.13\\
     \midrule
     \multirow{5}{*}{$\mathcal{V}(a)$}&Majority& 89.30& 31.45\\
    & Graph Prop & 84.62& 38.82\\
    &3-nn& 85.62& 38.45\\
    &Aspect-Level& 87.96&  48.06\\
    & Frame-Level & 87.96 & 48.06\\
     \midrule
     \multirow{5}{*}{$\mathcal{S}(t)$}&Majority& 71.91&  27.89\\
    & Graph Prop & 69.90 & 55.57\\
    &3-nn& 72.91& 59.26\\
    &Aspect-Level& 81.61&  72.85\\
    & Frame-Level & 81.61 & 72.85\\
     \midrule
     \multirow{5}{*}{$\mathcal{S}(a)$}&Majority& 50.84& 22.47\\
    &Graph Prop & 48.83  & 35.40\\
    &3-nn& 54.85& 45.51\\
    &Aspect-Level& 61.54&  53.88\\
    & Frame-Level & 61.54 & 53.88\\
    \bottomrule
    \end{tabular}
    }
    \caption{Detailed breakdown of results on the development set using accuracy and average F1 over the three class labels (+,-,=). }
    \label{results:dev}
\end{table}
\begin{table}[tb]
    \centering
    \begin{tabular}{lcc}
\toprule{\bf Algorithm} & {\bf Acc.}  & {\bf Avg F$_1$}\\ 
 \midrule
 Graph Prop & 58.81 & 41.46\\
 3-nn & 63.71 & 47.30\\
 \midrule
 Aspect-Level& 67.93 & 53.17\\
 Frame-Level & 68.26 & 53.50\\
 \bottomrule
    \end{tabular}
    \caption{Performance on the test set. Results are averaged over the different aspects.}
    \label{results:test}
\end{table}

\para{Results}
As shown in Table~\ref{results:dev}, aspect-level and frame-level models consistently outperform all three baselines --- \textsc{Majority, 3-NN, Graph Prop} in the development set across the different types of relationships. In particular, the improved F1 scores show that these models are able to perform better across all three classes of labels even in the most skewed cases.  The frame-level model also frequently improves the F1 scores of the labels from what they were in the aspect-level model.
The summarized comparison of the classifiers' performance test set is shown in Table~\ref{results:test}.  As with the development set, aspect-level and frame-level are both able to outperform the baselines. 
Furthermore, the frame-level formulation is able to make improvement over the results of the aspect-level classification, indicating that the modelling of inter-dependencies between relationships did help correct some of the mistakes made.


One point of interest about the frame-level results is whether the learned weights over the consistency factors match our initial intuitions about inter-dependencies between relationships. The weights learned in our algorithm do tell us something interesting about the degree to which these inter-dependencies are actually found in our data.

\begin{figure}[tb]
    \centering
    \begin{subfigure}[H]{.2\textwidth}
        \includegraphics[width=\textwidth]{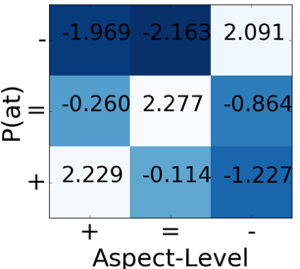}
        \caption{$\theta_{\mathcal{P}(a \rightarrow t)}$ }
        \label{weights:emb}
    \end{subfigure}
    \begin{subfigure}[H]{.2\textwidth}
        \includegraphics[width=\textwidth]{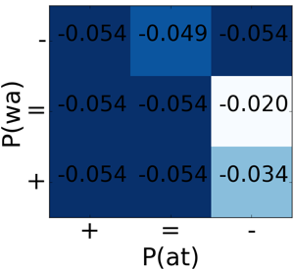}
        \caption{$\theta_{PT}$, $\mathcal{P}(w \rightarrow t)$: $-$}
        \label{weights:neg}
    \end{subfigure}
    \vspace*{3mm}

    \begin{subfigure}[H]{.2\textwidth}
        \includegraphics[width=\textwidth]{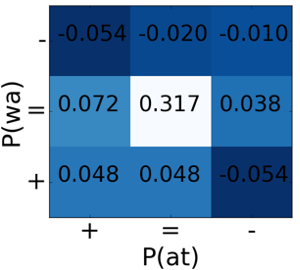}
        \caption{ $\theta_{PT}$, $\mathcal{P} (w \rightarrow t)$: $=$}
        \label{weights:neu}
    \end{subfigure}
    \begin{subfigure}[H]{.2\textwidth}
        \includegraphics[width=\textwidth]{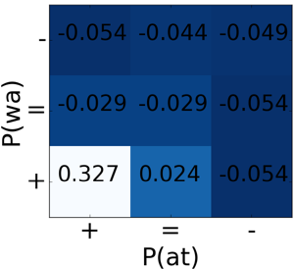}
        \caption{ $\theta_{PT}$, $\mathcal{P} (w \rightarrow t)$: $+$}
        \label{weights:pos}
    \end{subfigure}
    \vspace*{-2mm}
    \caption{\small Learned weights, $\theta$, of embedding factor for the perspective of agent to theme (\ref{weights:emb}) and the weights the perspective triad (PT) factor (\ref{weights:neg}-\ref{weights:pos}).  Lighter shades are for weights that are more positive, whereas dark blue is more negative.}
    \vspace*{-5mm}
    \label{weights}
\end{figure}

We show the heat maps for some of the learned weights in Figure~\ref{weights}. In ~\ref{weights:emb}, we show the weights of one of the embedding factors, and how the node's polarities are more strongly weighted when they match the aspect-level output. In the rest of the figure, we show the weights for the other perspective relationships when $\mathcal{P} (w \rightarrow t)$ is negative (\ref{weights:neg}), neutral (\ref{weights:neu}), and positive (\ref{weights:pos}), respectively.  Based on the expected interdependencies, when $\mathcal{P} (w \rightarrow t): -$, the model should favor $\mathcal{P} (w \rightarrow a) \neq \mathcal{P} (a \rightarrow t)$ and when $\mathcal{P} (w \rightarrow t): +$, the model should favor $\mathcal{P} (w \rightarrow a) = \mathcal{P} (a \rightarrow t)$.  Our model does, in fact, learn a similar trend, with slightly higher weights along these two diagonals in the maps \ref{weights:neg} and \ref{weights:pos}.  Interestingly, when $\mathcal{P} (w \rightarrow t)$ is neutral, weights slightly prefer for the other two perspectives to resemble one another, but with highest weights being when other perspectives are also neutral.

%% file: analysis.tex
\cut{
\begin{table}[tb]
\begin{center}
\small
\begin{tabular}{cc}
\toprule
\bf Left & \bf Right \\
\cmidrule(r){1-1}
\cmidrule(l){2-2}
huffingtonpost.com    &  foxnews.com  \\
slate.com    &  nypost.com  \\
thinkprogress.org & washingtontimes.com    \\
newsweek.com    &  theamericanconservative.com  \\
politico.com    &  weeklystandard.com  \\
washingtonpost.com    &  nationalreview.com  \\
nytimes.com    &   townhall.com \\
latimes.com    &   lifesitenews.com \\
newyorker.com    &   nationalrighttolifenews.org \\
salon.com    &   breitbart.com \\
theguardian.com    &   wnd.com \\
thedailybeast.com    &  city-journal.org  \\
opednews.com    &   thehill.com \\
rhrealitycheck.org    &  humanevents.com  \\
npr.org    &   theblaze.com \\
\bottomrule
\end{tabular}
\end{center}
\caption{Online news sources selected for analysis.}
\label{tab:analysis:sources}
\end{table}
}

\subsection{Analysis of a Large News Corpus}
\label{ssec:analysis}


Using the connotation frame, we present measured implied sentiment in online journalism.

\paragraph{Data}
%
From the Stream Corpus~\cite{streamcorpus14}, we select $70$ million news articles. 
We extract subject-verb-object relations for this subset using the direct dependencies between noun phrases and verbs as identified by the BBN Serif system, 
obtaining $1.2$ billion unique tuples of the form \emph{(url,subject,verb,object,count)}. 
 We also extract tuples from news articles from the Annotated English Gigaword Corpus~\cite{gigaword}, which contains nearly 10 million articles, resulting in an additional 120 million unique tuples.

\begin{figure}[tb]
\centering
\hspace*{-6mm}
\includegraphics[width=1.1\columnwidth,clip,trim=22 22 30 57]{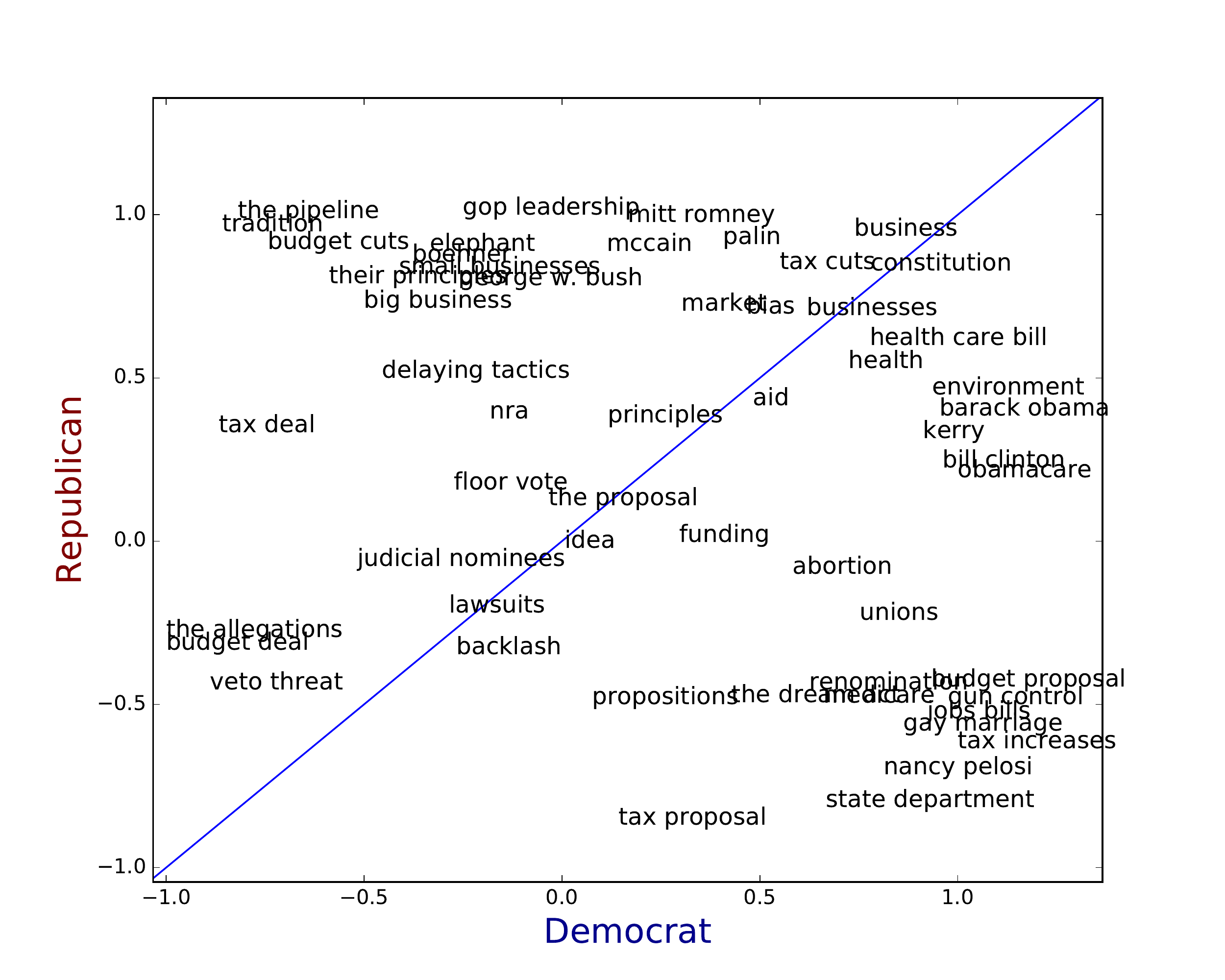}
\caption{\small Average sentiment of Democrats and Republicans (as agents) to selected nouns (as their themes), aggregated over a large corpus using the learned lexicon (\S\ref{ssec:perspective}). The line indicates identical sentiments, i.e. Republicans are more positive towards the nouns that are above the line.}
\label{fig:analysis:subobj}
\end{figure}

\paragraph{Estimating Entity Polarities}
Using connotation frames, we can also measure entity-to-entity sentiment at a large scale. 
Figure~\ref{fig:analysis:subobj}, for example, presents the polarity of entities ``Democrats'' and ``Republicans'' towards a selected set of nouns, by computing the average estimated $\mathcal{P}(a\rightarrow t)$ polarity (using our frame-level output) over triples where one of these entities appears as part of the phrase in the agent role (e.g. ``Democrats'' or ``Republican party'').
\hannahedit{Apart from nouns that both entities are positive~(``business'', ``constitution'') or negative~(``the allegations'',``veto threat'') towards, we can also see interesting examples in which Democrats feel more positively (below the line: ``nancy pelosi'', ``unions'', ``gun control'', etc.) and ones where Republicans feel more positive~(``the pipeline'', ``gop leadership'', ``budget cuts'', etc.). Also, both entities are neutral towards ``idea'' and ``the proposal'', which probably owes to the fact that ideas or proposals can be good or bad for either entity depending on the context. }

%% file: related.tex
\section{Related Work}

Most prior work on sentiment lexicons focused on the overall polarity of words without taking into account their semantic arguments \cite{wilson2005recognizing,sentiwordnet3,wiebe2005annotating,Velikovich:2010,kaji2007building,kamps2004using,taka05,alina06}. Several recent studies began exploring more specific and nuanced aspects of sentiment such as connotation~\cite{feng2013connotation}, good and bad effects~\cite{choi-wiebe:2014:EMNLP2014}, and evoked sentiment~\cite{mohammad2010emotions}. Drawing inspirations from them, we present connotation frames as a unifying representation framework to encode the rich dimensions of implied sentiment, presupposed value judgements, and effect evaluation, and propose a factor graph formulation that captures the interplay among different types of connotation relations.  

Goyal et al.~\shortcite{Goyal:2010:APP:1870658.1870666,daume10plotunits} investigated how characters (protagonists, villains, victims) in children's stories are affected by certain predicates, which is related to the effect relations studied in this work. 
%
%
While \newcite{klenner2014verb} similarly investigated the relation between the polarity of the verbs and arguments, our work introduces new perspective types and proposes a unified representation and inference model. \newcite{wiegand2015} also looked at perspective-based relationships induced by verb predicates with a focus on opinion roles. Building on this concept, our framework also incorporates information about the perspectives' polarities as well as information about other typed relations.
There have been growing interests for modeling framing \cite{Greene:2009:MWS:1620754.1620827,hasan2013frame}, 
biased language \cite{recasens2013linguistic} and ideology detection \cite{Yano:2010:SLB:1866696.1866719}. All these tasks are relatively less studied, and we hope our connotation frame lexicon will be useful for them.


Sentiment inference rules have been explored by the recent work of 
\newcite{DBLP:journals/corr/WiebeD14} and \newcite{deng2014sentiment}. The focus of their work was on general inference rules that are not predicate-specific. 
In contrast, our work focuses on the notion that connotative polarities can be determined directly from the predicate, rather than partial knowledge of the arguments or the context in which it is being used.  
In brief, we make a novel conceptual connection between inferred sentiments and frame semantics, organized as connotation frames, and present a unified model that integrates different aspects of the connotation frames.
%
%

Finally, in a broader sense, what we study as connotation frames draws a connection to schema and script theory  \cite{schank1975scripts}. Unlike  prior work that focused on 
directly observable actions~\cite{Chambers:2009:ULN:1690219.1690231,
fermann:EACL2014,bethard2008building}, we focus on implied sentiments that are framed by predicate verbs. 

%% file: conclusions.tex
\section{Conclusion}

In this paper, we presented a novel system of connotative frames that define a set of implied sentiment and presupposed facts for a predicate. Our work also empirically explores different methods of inducing and modelling these connotation frames, incorporating the interplay between relations within frames.
%
Our work suggests new research avenues on learning connotation frames, and their applications to deeper understanding of social and political discourse. 
All the learned connotation frames and annotations are available at \url{http://homes.cs.washington.edu/~hrashkin/connframe.html}.